\documentclass[10pt, lettersize,journal]{IEEEtran}
\usepackage{amsmath,amsfonts}
\usepackage{algorithmic}
\usepackage{algorithm}
\usepackage{array}
\usepackage[caption=false,font=normalsize,labelfont=sf,textfont=sf]{subfig}
\usepackage{textcomp}
\usepackage{stfloats}
\usepackage{xcolor}
\usepackage{url}
\usepackage{verbatim}
\usepackage{graphicx}
\usepackage{cite}
\usepackage{booktabs}
\usepackage{multirow}
\usepackage[normalem]{ulem}  % 提供 \sout 删除线命令

\begin{document}

% \title{Only Train Once: Uncertainty-Aware Self-Supervised Learning for Face Authenticity Detection}
\title{Only Train Once: Uncertainty-Aware One-Class Learning for Face Authenticity Detection}

\author{Qingchao Jiang \IEEEmembership{Senior Member, IEEE}, Zhenxuan Hou, Zhiying Zhu, Zhenxing Qian, \IEEEmembership{Senior Member, IEEE}, Xinpeng Zhang, \IEEEmembership{Senior Member, IEEE} , Zaiwang Gu
        % <-this % stops a space

\thanks{This work was supported by the National Natural Science Foundation of China under Grant 62402182, 62322309, 62572125, the Natural Science Foundation of Shanghai under Grant 25ZR1401019, the Shanghai Explorer Program under Grant 24TS1411700, and the Fundamental Research Funds for the Central Universities under Grant B260201918. }

\thanks{Qingchao Jiang, Zhenxuan Hou and Zhiying Zhu are with the School of Information Science and Engineering, East China University of Science and Technology, Shanghai 200237, China(e-mail: qchjiang@ecust.edu.cn; Y80240154@mail.ecust.edu.cn; zyzhu@ecust.edu.cn)}

\thanks{Zhenxing Qian and Xinpeng Zhang are with College of Computer Science and Artificial Intelligence, Fudan University, Shanghai 200433, China(e-mail: zxqian@fudan.edu.cn; zhangxinpeng@fudan.edu.cn).}
\thanks{Zaiwang Gu is with the Institute of Advanced Intelligence and Computing (IAIC), Agency for Science, Technology and Research (A*STAR), Singapore.(e-mail: 
guzw@a-star.edu.sg; ).}% <-this % stops a space
\thanks{Corresponding author: Zhiying Zhu.}
\thanks{This paper has supplementary downloadable material available at http://ieeexplore.ieee.org, provided by the authors.  }
}

\markboth{Journal of \LaTeX\ Class Files,~Vol.~14, No.~8, August~2021}%
{Shell \MakeLowercase{\textit{et al.}}: A Sample Article Using IEEEtran.cls for IEEE Journals}

\IEEEpubid{0000--0000/00\$00.00~\copyright~2021 IEEE}

\maketitle

\begin{abstract}

 % The prevalence of AI-generated images has sparked significant concerns regarding the potential for malicious exploitation of image generation techniques. 
 The rapid evolution of generative paradigms has enabled the creation of highly realistic imagery, which escalates the risks of identity fraud and the dissemination of disinformation. Most existing approaches frame face forgery detection as a fully supervised binary classification problem. Consequently, these models typically exhibit significant performance decay when tasked with detecting forgeries from previously unseen generative paradigms. Furthermore, these methods focus exclusively on either DeepFakes or fully synthesized faces, thereby failing to provide a generalized framework for universal face forgery detection. In this paper, we address this challenge by introducing FADNet (Face Authenticity Detector Net), 
 % a self-supervised framework that 
 which reformulates face forgery detection as a one-class classification (OCC) task. By training exclusively on authentic facial data to capture their intrinsic representations, FADNet flags any image whose feature embedding deviates significantly from the learned distribution of real faces as a forgery. In this way, FADNet requires only a single training process without retraining for each specific forgery method. The framework incorporates Evidential Deep Learning (EDL) to quantify predictive uncertainty and utilizes a plug-and-play pseudo-forgery image generator (PFIG) to tighten decision boundaries around authentic data. Extensive experimental evaluations on the DF40 and ASFD benchmarks demonstrate that FADNet achieves superior performance and generalization capabilities. Specifically, FADNet substantially outperforms existing state-of-the-art (SOTA) methods, yielding a remarkable average accuracy of 96.63\% and an average precision of 98.83\%. Our code is available at
\url{https://github.com/hzx111621/Only-Train-Once}.
 % These results indicate that focusing on the characterization of authentic facial features rather than model-specific fingerprints is a more viable and sustainable strategy for long-term forgery detection. Because it is impractical for training sets to encompass the entire landscape of potential forgeries, there is an urgent need for a detection strategy that is decoupled from the architectural dependencies of specific generative models.
\end{abstract}


\section{Introduction}
\label{sec:sec1}
Recent breakthroughs in deep learning have significantly lowered the barrier to generating high-fidelity face forgeries~\cite{9115874, 10444967, 10400952,10814723,lin2025learningrepresentationsynergyinvariances}. In particular, Deepfake technology~\cite{TOLOSANA2020131,wang2024deepfake}, which utilizes deep neural networks to perform face swapping, has emerged as a major tool for digital manipulation. Concurrently, image generation has undergone remarkable progress over the past few years. A succession of generative paradigms has emerged, encompassing autoregressive models~\cite{NIPS2016_ddeebdee},  normalizing flows~\cite{NEURIPS2018_d139db6a}, generative adversarial networks (GANs)~\cite{TREVISANDESOUZA202313}, and diffusion models (DMs)~\cite{10081412}. Images generated by these models exhibit high fidelity, rendering them nearly indistinguishable to the human eye. These technologies have been widely utilized in fields such as virtual reality and digital entertainment~\cite{10979443}. However, these advancements are also being exploited for illegal purposes, including identity fraud and the dissemination of disinformation~\cite{10401723}. Therefore, distinguishing the authenticity of facial imagery has become an imperative task.
\begin{figure}
    \centering
    \includegraphics[width=0.8\linewidth]{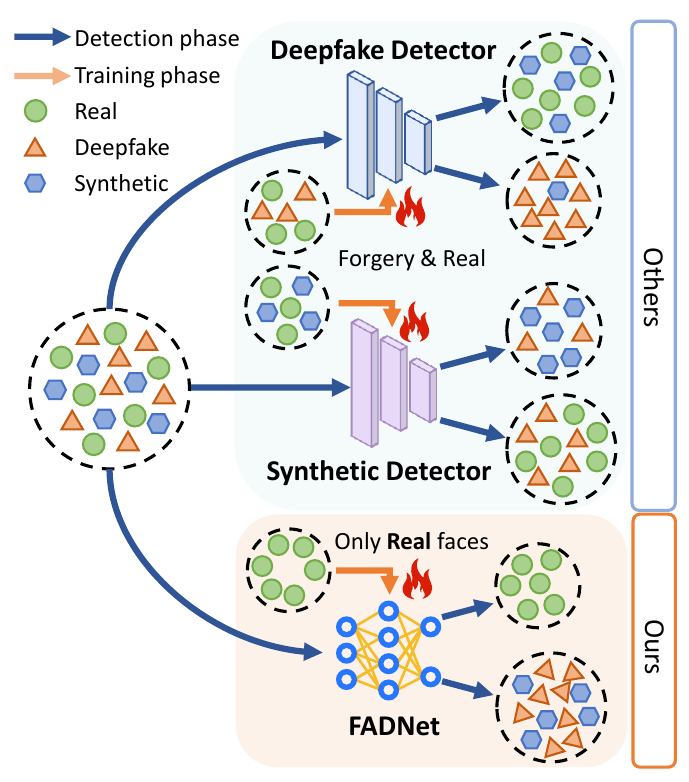}
    \caption{Comparison between existing method and ours. In the training phase, unlike existing supervised methods that rely on large-scale collections of forged faces, our FADNet requires only real face images for training. In the detection phase, the \textbf{Deepfake Detector} excels at identifying Deepfakes but generalizes poorly to synthetic faces, the \textbf{Synthetic Detector}, conversely, excels on synthetic faces but is less effective against Deepfakes. \textbf{Our method} effectively identifies both Deepfakes and synthetic faces within a unified detection framework.}
    \label{fig:fig1}
\end{figure}
% Comparison between our method and existing methods. Unlike existing supervised methods that require large-scale collections of forged faces, our FADNet adopts a self-supervised approach, necessitating only real faces for training.

Recently, a lot of detection methodologies~\cite{11015656,zhu2021face,wang2023dire,11209901,11098842,11145290,wang2020cnn, 11329196, MIR-2025-09-551, 10814697,Lai_2026_CVPR} have been introduced to identify image forgeries, encompassing deepfakes or fully synthetic facial imagery. \IEEEpubidadjcol Existing approaches typically frame face forgery detection as a binary classification problem.  These methods predominantly adopt a fully supervised learning paradigm, where models are trained on real photographic images alongside forgeries produced by one or more specific generative models, and are subsequently evaluated on imagery from previously unseen sources. For instance, reference~\cite{wang2020cnn} has demonstrated that images synthesized by CNN-based models exhibit systematic and detectable artifacts. Notably, a classifier trained solely on imagery from a single generator (e.g., ProGAN) can exhibit a surprising degree of generalization, effectively discerning authentic images from those produced by a wide range of unseen CNN architectures. Reference~\cite{11209901} observed that images synthesized by a specific generative method tend to yield significantly lower reconstruction errors when processed by a matching model than by a mismatched one. Leveraging this insight, the authors proposed a multi-reconstruction detection framework that performs inverse mapping of input face images through an ensemble of pre-trained generative models and distinguishes authentic faces from synthetic ones generated by both GANs and diffusion models by analyzing the resulting reconstruction discrepancies. Because limited training datasets cannot comprehensively cover the full spectrum of potential forgeries, generalization has become a fundamental requirement for modern detectors. Regrettably, the rapid evolution of generative models causes most existing detectors to suffer from substantial performance degradation when confronted with novel architectures. Given that generative paradigms will continue to evolve alongside advancements in synthesis technology, relying on model-specific fingerprints or reconstruction error for face forgery detection is no longer a viable or sustainable strategy.

To solve this issue, in this paper, we try to decouple the detection framework from the architectural dependencies of forgery models. The differences between our method and the others are shown in Fig.~\ref{fig:fig1}. We reformulate face forgery detection as an OCC task, in which manipulated faces are treated as anomalous samples. Consequently, the objective shifts from detecting forgeries to Face Authenticity Detection. Building upon this foundation, we propose FADNet.
% By leveraging a self-supervised paradigm, the framework learns representations exclusively from pristine faces, thereby enabling the effective differentiation of forged samples. 
By learning representations exclusively from pristine facial imagery, the framework is capable of effectively discriminating forged samples. Meanwhile, the model only needs to be trained once and does not require retraining when encountering newly emerging forgery methods. Furthermore, we incorporate EDL to quantify the model's predictive uncertainty by parameterizing a Dirichlet distribution. During training, authentic images serve as in-distribution (ID) data. Utilizing the inherent properties of evidential learning, the model naturally assigns higher uncertainty to out-of-distribution (OOD) fake images, thereby facilitating a natural separation between real and forged content. To bolster the discriminative power of FADNet, we also introduce a plug-and-play pseudo forgery image module that transforms pristine faces into pseudo forgery images. These samples act as OOD data during training, guiding the model to tighten the decision boundaries around the authentic data distribution.

% We proposed FADNet to capturing the intrinsic representations of authentic faces. By adopting a self-supervised paradigm, we reformulate face forgery detection as a OCC task, where fake face are treated as anomalous samples. Any image whose feature embedding deviates significantly from the learned distribution of real faces is flagged as a forgery. 

Comprehensive experimental results demonstrate that our FADNet effectively identifies both DeepFakes and fully synthesized facial imagery, consistently outperforming SOTA benchmarks by a significant margin. Notably, our method not only achieves superior performance on existing forgery methods but also maintains good generalization capabilities against the newest synthesis paradigms. By integrating EDL, the model provides calibrated uncertainty scores and significantly mitigates model overconfidence, thereby enhancing the overall reliability of the detection process. Our contributions can be summarized as follows:
\begin{enumerate}
    \item We reformulate the face forgery detection problem by shifting the focus toward facial authenticity, thereby framing it as an OCC task. 
    \item We propose FADNet, a novel framework that integrates one-class learning and Evidential Deep Learning for face authenticity detection.
  
    \item Comprehensive evaluations show that our approach effectively counters diverse facial forgeries. Moreover, it enables the detection of faces from unknown forgery paradigms after only one training session.

\end{enumerate}

The paper is structured as follows. Section~\ref{sec:sec2} presents an overview of related works. Section~\ref{sec:sec3} details the proposed
methodology. Section~\ref{sec:sec4} presents experimental results and comprehensive analysis. Finally, Section~\ref{sec:sec5} concludes the paper.

\section{related work}
\label{sec:sec2}
This section provides a brief overview of face forgery techniques, face forgery detection methodologies, and Evidential Deep Learning.
\subsection{Face Forgery}
Recent advances in face forgery can be broadly categorized into two domains: Deepfakes~\cite{pei2024deepfake} and synthetic faces~\cite{9667066}. Deepfakes primarily leverage deep learning frameworks, such as GANs and Autoencoders (AEs), to model the underlying distribution of real facial data, thereby enabling high-fidelity manipulation of facial features. Research into deepfake generation is currently concentrated in four mainstream directions: 1) Face Swap~\cite{xu2022region}, Focused on interchanging identities between two individuals; 2) Face Reenactment~\cite{hsu2022dual}, Emphasizing the transfer of motions and poses from a source subject to a target; 3) Talking Face Generation~\cite{zhou2019talking}, Centered on achieving seamless synchronization between lip movements and textual or acoustic content; 4) Face Attribute Editing~\cite{pang2023dpe}, Aiming for the precise modification of specific facial characteristics within a target image. Recent studies have moved beyond task-specific face manipulation toward unified frameworks. For example, FaceAnime\cite{tu2021image} leverages 3D facial dynamics for controllable face reenactment. Both UniFace\cite{xu2022designing} and UniFace++\cite{xu2025uniface++} achieve unified face reenactment and swapping, with the former using an end-to-end disentangled framework and the latter incorporating reassembled 3D facial priors with diffusion-based reconstruction.
% Recently, Deepfakes have evolved toward unified face manipulation frameworks integrating reenactment and swapping. For instance, DiffSFSR\cite{liu2024towards} achieves granular identity and expression control via diffusion models, while UniFace++\cite{xu2025uniface++} leverages explicit 3D priors for stable and geometrically controllable synthesis.

In contrast, synthetic face generation aims to synthesize photorealistic facial images without relying on specific identity references. This line of research mainly focuses on learning the global data distribution of real faces and generating samples through latent space sampling and decoding. Representative methods are primarily based on GANs and DMs, which have demonstrated remarkable capability in producing high-resolution and semantically consistent facial imagery. By leveraging progressive training strategies, hierarchical feature modulation, and iterative denoising processes, these models achieve fine-grained control over facial attributes and structural details.

\subsection{Face Forgery Detection}
Recent research has primarily focused on two avenues to enhance the generalization of face forgery detection: tampering traces (including generative model fingerprints) and reconstruction discrepancy.
Early studies focused on spatial statistical inconsistencies, such as pixel-level blending artifacts, illumination inconsistencies, or anomalous color space distributions~\cite{li2020face}.  Simultaneously, researchers observed that upsampling operations within generative networks inevitably introduce periodic spikes in the frequency domain~\cite{pmlr-v119-frank20a}, creating unique "spectral fingerprints." Reference~\cite{yu2019attributing} further demonstrated that each GAN model possesses a specific intrinsic fingerprint that can be leveraged for image attribution and forensics. Some studies have further explored noise-based and temporal semantic cues. \cite{wang2023noise} proposed NoiseDF, which detects deepfakes by extracting forensic noise from facial and background regions and modeling the discrepancies between them. However, as generative algorithms increasingly prioritize perceptual quality, post-processing techniques (smoothing and resampling) can easily obscure these subtle statistical traces. This results in a significant degradation of detection performance in cross-dataset scenarios.

To overcome the limitations of fragile traces, a class of detection frameworks based on reconstruction discrepancy has emerged. Instead of searching for local defects, these methods evaluate the authenticity of an image by verifying its compatibility with the generative model manifold. The core hypothesis is that synthetic images exhibit high reconstruction fidelity within their corresponding generative space, whereas real images cannot be perfectly reconstructed by a generative model. For images generated by Diffusion Models, reference~\cite{wang2023dire} proposed DIRE, which employs a process of inversion followed by reconstruction. By employing the reconstruction residual as a discriminative criterion, DIRE facilitates the effective detection of synthetic faces. Furthermore, reference~\cite{11209901} proposed Model Discrepancy Learning , which constructs multiple reconstruction branches to compare the stability of an image across different reconstruction paths. This allows for the identification of subtle deviations on the generative manifold without the need for explicit fingerprints. 

In addition, FractalForensics\cite{wang2025fractalforensics} adopts a semi-fragile fractal watermarking strategy for active deepfake detection and can further localize tampered regions. \cite{wang2026word} proposed the WAFL paradigm, which simplifies temporal localization in audio-visual Deepfakes into a word-level discrete binary classification task, enabling efficient and accurate forgery localization.
However, existing detection methods often suffer from significant performance degradation when encountering unseen generative architectures. Although recent methods such as OC-FakeDect\cite{khalid2020oc} and SeeABLE\cite{10376925} have relaxed the dependency on forgery faces, they cannot maintain high performance on both Deepfakes and synthetic faces simultaneously.

% And they cannot maintain high performance on both Deepfakes and synthetic faces simultaneously.
\subsection{Evidential Deep Learning}
Evidential Deep Learning estimates the uncertainty in classification predictions by modeling the Dirichlet distribution of class probabilities. Unlike Bayesian Neural Networks (BNNs)~\cite{mackay1992practical}, which indirectly infer predictive uncertainty via weight distributions, EDL explicitly models uncertainty based on Subjective Logic theory and the Dempster-Shafer Theory (DST)~\cite{dempster1968generalization} of evidence.

For a given input $x_i$ and $K$ classes, the neural network outputs an evidence vector ${e}_i = [e_{i1}, e_{i2}, \ldots, e_{iK}]^T$, where each element represents the model's support for the corresponding class. To ensure the non-negativity of the evidence, a non-negative activation function is applied to the network's output, such as ReLU. The evidence vector ${e}_i$ is subsequently utilized to define the parameters ${\alpha}_i$ of the Dirichlet distribution, formulated as:
\begin{equation}
    \alpha_{ik} = e_{ik} + 1,
    \label{eq:alpha}
\end{equation}
where $k$ denotes the index of each class. %The resulting Dirichlet distribution $p_i \sim \text{Dir}(\boldsymbol{\alpha}_i)$ (where $p_i$ represents the class probabilities) allows for the modeling of predictive uncertainty.

According to Subjective Logic theory, the uncertainty $u_i$ associated with the input $x_i$ is defined as:
\begin{equation}
    u_i = \frac{K}{\sum_{k=1}^{K} \alpha_{ik}},
    \label{eq:uncertainty}
\end{equation}
where $\sum_{k=1}^{K} \alpha_{ik}$ represents the sum of Dirichlet parameters across all classes. Here, $u_i$ is inversely correlated with the accumulated evidence. A scenario with zero total evidence yields maximum uncertainty ($u_i=1$), signifying complete ignorance of the model regarding the input.
\section{proposed method}
\label{sec:sec3}
This section elucidates the technical details of the proposed FADNet. We first establish a formal problem definition in Section~\ref{sec:subsec1}. The architectural design of FADNet is detailed in Section~\ref{sec:subsec2}. Subsequently, Section~\ref{sec:subsec3} introduces a plug-and-play module, PFIG. Section~\ref{sec:subsec4} provides a comprehensive description of the training and optimization. Finally, Section~\ref{subsec:inference} introduces the decision strategy of FADNet.
\subsection{Problem Statement}
\label{sec:subsec1}

Given an input facial image $I \in \mathbb{R}^{H \times W \times 3}$, the objective of face forgery detection is to discern whether $I$ is an authentic photo or a manipulated forgery. A more formidable challenge lies in developing a detector capable of generalizing across a diverse set of forgery methods $\mathcal{M} = \{M_1, M_2, \ldots, M_K\}$ encompassing techniques such as Face2Face and FaceSwap, as well as generative models like StyleGAN and Stable Diffusion.

Formally, we aim to learn a mapping function $f$:
\begin{equation}
f: I \rightarrow (y, u) \in \{0,1\} \times [0,1],
\end{equation}
subject to the following objectives:
\begin{equation}
\forall I_r \in \mathcal{D}_{\text{real}}: f(I_r) = (1, u_{\text{low}}),
\end{equation}
\begin{equation}
\forall I_f \in \mathcal{D}_{\text{forgery}}: f(I_f) = (0, u_{\text{high}}),
\end{equation}
where $y=1$ and $y=0$ denote real and forged images, respectively. $u_{low}$ and $u_{high}$ denote low and high uncertainty, respectively.  Here, $\mathcal{D}_{\text{real}}$ represents the dataset of real faces, and $\mathcal{D}_{\text{forgery}} = \mathcal{D}_{\text{deepfake}} \cup \mathcal{D}_{\text{synthetic}}$ constitutes the set of forgeries, which includes:
\begin{enumerate}
    \item $\mathcal{D}_{\text{deepfake}}$: Deepfake instances synthesized via various manipulation modalities.
    \item $\mathcal{D}_{\text{synthetic}}$: Faces fully synthesized by a generative model.
\end{enumerate}

The term $u \in [0,1]$ represents the uncertainty of the model. The constraint $u_{\text{low}} \ll u_{\text{high}}$ ensures that the model maintains minimal uncertainty for real images while yielding high uncertainty for forged ones.

Unlike fully supervised binary classification, our approach adopts a  one-class learning paradigm, trained exclusively on authentic face images. Specifically, we focus on characterizing the feature distribution of real facial data, identifying samples that significantly deviate from this distribution as forgeries. This design facilitates superior generalization to unseen forgery techniques during inference, as it obviates the need to explicitly model the exhaustive space of all possible manipulation and generation patterns.

\subsection{FADNet Architecture}
\label{sec:subsec2}

\begin{figure*}
    \centering
    \includegraphics[width=0.9\linewidth]{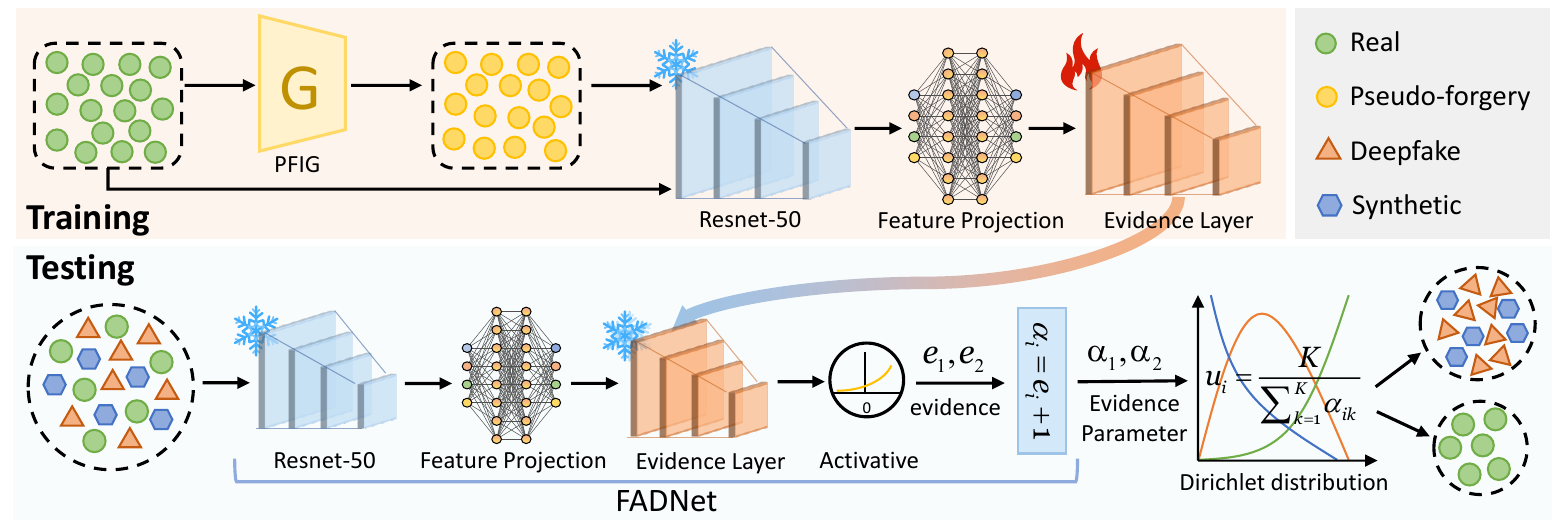}
    \caption{The overall architecture of FADNet. During the training phase, a PFIG is utilized to create pseudo-forgery samples, which are used in conjunction with real images to train FADNet. Once the training is complete, the model parameters are frozen. In the inference stage, authentic images yield low uncertainty values, whereas all other inputs generate high uncertainty values, thereby facilitating effective differentiation.}
    \label{fig:fig2}
\end{figure*}

The overall architecture of FADNet is shown in Fig.~\ref{fig:fig2}. We employ a pre-trained ResNet-50 as the backbone for feature extraction. For an input image $I \in \mathbb{R}^{224\times224\times3}$, the ResNet-50 backbone yields a 2048-dimensional raw feature vector $\mathbf{f}_{\text{raw}}$. To project these features into a more compact and discriminative embedding space, we utilize a two-layer Multi-Layer Perceptron (MLP):
\begin{equation}
\mathbf{f} = \mathbf{W}_2 \cdot \text{ReLU}(BN(\mathbf{W}_1 \mathbf{f}_{\text{raw}}))
\end{equation}
where $\mathbf{W}_1$ and $\mathbf{W}_2$ denote the learnable weight matrices and $\text{BN}$ represents Batch Normalization. The resulting feature vector $\mathbf{f} \in \mathbb{R}^{512}$ is subsequently processed via $L_2$ normalization to ensure stable optimization.

The normalized feature $\mathbf{f}$ is fed into an evidential layer that parameterizes the concentration parameters of a Dirichlet distribution:

\begin{equation}
e=\text{Softplus}(\text{MLP}(\mathbf{f})), 
\alpha = e + 1 
\end{equation}
where ${\alpha} = [\alpha_1, \alpha_2]^\top$ represents the evidence strength for class $k \in \{0, 1\}$ (forgery vs. authentic). The addition of 1 ensures that the Dirichlet parameters satisfy $\alpha_k \ge 1$, consistent with Subjective Logic theory.

We derive the predictive uncertainty $u(I)$ based on the Dirichlet distribution:
\begin{equation}
u(I) = \frac{K}{S}, \quad S = \sum_{k=1}^{K} \alpha_k
\end{equation}
where $K=2$ is the number of classes and $S$ is the total Dirichlet strength. A higher value of $u(I)$ indicates a higher degree of vacuity, signifying that the model is more uncertain about the photo.

\subsection{Pseudo-forgery Image Generator}
\label{sec:subsec3}
As modern generative models become capable of synthesizing highly realistic facial images that are nearly indistinguishable to the human eye, detectors trained solely on authentic samples often lack sufficient discriminative power when encountering fully synthetic forgeries. To address this issue, we designed the PFIG module. This module constructs a series of representative proxy forgeries by applying specific perturbations to real faces. By treating these proxy forgeries as OOD samples during training, the model is forced to tighten its decision boundaries to more closely encompass the real manifold. Unlike conventional data augmentation, which aims to preserve semantic consistency, PFIG deliberately perturbs images to push them away from the authentic face manifold, thereby simulating potential forgery artifacts. Unlike SeeABLE, which applies subtle spatial or frequency perturbations to a single grid region and supervises the prediction of their locations and types, PFIG jointly introduces local, multi-region, and global anomalies across frequency, spatial-structural, chromatic-textural, and compression domains.
Specifically, we design four types of perturbations that effectively assist FADNet in learning a more compact manifold representation for pristine images:
%Based on different feature dimensions, we categorize these perturbations into four types:

\subsubsection{\textbf{Frequency Domain Perturbations}}
\begin{itemize}
    \item Spectral Disruptions: Injects periodic peaks into high-frequency regions to disrupt the statistical consistency of the power spectrum distribution.
\end{itemize}
\subsubsection{\textbf{Spatial and Structural Inconsistencies}}
\begin{itemize}
    \item Local Blending: Randomly blends spatial regions to simulate topological discontinuities inherent in image splicing and synthesis processes.
    \item Spatial Sharpness Inconsistency: Mixes blurred and sharpened patches to simulate non-uniformity in clarity across the facial manifold.
    \item Checkerboard Artifacts: Simulating the signature footprints introduced by deconvolutional (transposed convolutional) layers.
\end{itemize}
\subsubsection{\textbf{Chromatic and Textural Statistics}}
\begin{itemize}
    \item Color Channel Shift: Introduces offsets within color channels to simulate anomalies in color space distribution and covariance mismatch.
    \item Texture Perturbation: Applies local sharpening, smoothing, or noise to simulate the limitations of generative models in fine-grained texture modeling.
    \item Structured Noise: Injects blocky noise patterns to simulate sensor-specific noise fingerprints or structural noise residuals produced by specific algorithms.
\end{itemize}
\subsubsection{\textbf{Compression}}
\begin{itemize}
    \item JPEG Compression: Simulates post-processing traces and information entropy loss typically found in real-world transmission scenarios.
\end{itemize}

\begin{figure*}
    \centering
    \includegraphics[width=0.9\linewidth]{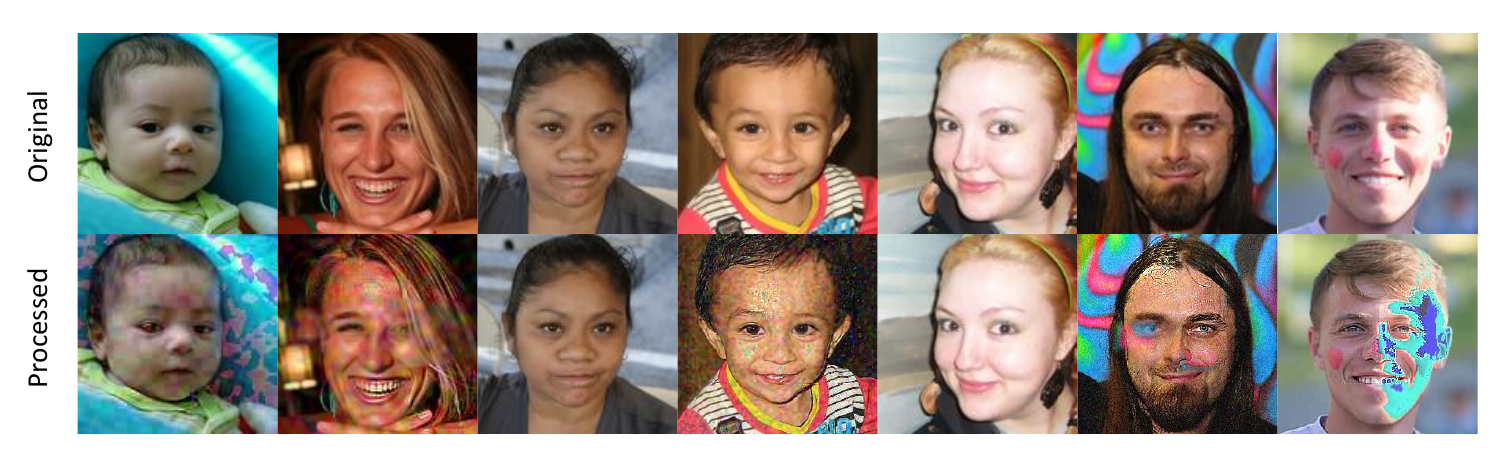}
    \caption{Comparison between pristine images and their pseudo-forgery counterparts. These pseudo-forgery samples are derived from authentic images through various transformations, exhibiting diverse visual discrepancies.}
    \label{fig:compare}
\end{figure*}

During the training phase, to ensure the diversity and robustness of the images, we randomly select a combination of 3 to 5 perturbations from the aforementioned set for each pristine image. Detailed descriptions and perturbation strengths are provided in the supplementary material. This stochastic augmentation strategy enables the model to learn a generalized boundary for forgeries without over-fitting to any single manipulation type. A visual comparison between the pristine images and the resulting pseudo-forgery counterparts is illustrated in Fig. \ref{fig:compare}.

\subsection{Training and Optimization}
\label{sec:subsec4}
We design a multi-objective loss function to jointly optimize classification accuracy, feature compactness, and uncertainty distribution. The total loss $\mathcal{L}_{\text{total}}$ is defined as:
\begin{equation}
\mathcal{L}_{\text{total}} = \mathcal{L}_{{EDL}} + \mathcal{L}_{{unc}} + \lambda_c \mathcal{L}_{{compact}} 
\end{equation}

\subsubsection{Evidential Loss}
We employ a Dirichlet-based cross-entropy loss:
\begin{equation}
\mathcal{L}_{{EDL}} = \frac{1}{|\mathcal{B}|} \sum_{(I,y) \in \mathcal{B}} \sum_{k=0}^{1} \mathbb{I}[y=k] \left(\Psi(S) - \Psi(\alpha_k)\right)
\end{equation}
where $\Psi(\cdot)$ denotes the digamma function, $\mathcal{B}$ is the batch size. $S = \alpha_1 + \alpha_2$ is the total Dirichlet strength, and $\mathbb{I}[\cdot]$ is the indicator function. This loss encourages the model to accumulate evidence $\alpha_k$ for the correct class, effectively maximizing the concentration of the Dirichlet distribution toward the ground-truth label.

\subsubsection{Uncertainty Regularization}
To differentiate between authentic and forged samples in the feature space, we explicitly constrain their respective uncertainty distributions:
\begin{equation}
\mathcal{L}_{{unc}} =  \frac{1}{N_r} \sum_{i=1}^{N_r} u(I_r^{(i)}) + \frac{1}{N_p} \sum_{I \in \{I_p\}} \|{\alpha}(I) - 1\|_1
\end{equation}
where $N_r$ and $N_p$ denote the number of real and pseudo-forgery images, respectively. The first term penalizes high uncertainty in real images, compelling the model to yield high-confidence predictions for authentic images. The second term suppresses evidence accumulation for forged images, ensuring they maintain high uncertainty.

\subsubsection{Feature Compactness Loss}
To encourage real faces to form a cohesive cluster in the embedding space, we introduce a compactness loss (center loss):
\begin{equation}
\mathcal{L}_{{compact}} = \frac{1}{N_r} \sum_{i=1}^{N_r} \left\| \mathbf{f}(I_r^{(i)}) - \bar{\mathbf{f}}_r \right\|_2^2
\end{equation}
where $\bar{\mathbf{f}}_r = \frac{1}{N_r} \sum_{i=1}^{N_r} \mathbf{f}(I_r^{(i)})$ represents the feature centroid of the authentic images within the current batch. This loss minimizes intra-class variance for real faces, thereby enhancing their separability from forgeries. We empirically set $\lambda_c = 0.5$.

\subsection{Inference and Decision Strategy}
\label{subsec:inference}

During the inference phase, for a given input image $I$, the model produces the Dirichlet parameters ${\alpha}(I)$ and the corresponding predictive uncertainty $u(I) = 2 / (\alpha_1 + \alpha_2)$. We adopt an uncertainty-based thresholding strategy for classification:
\begin{equation}
y = \begin{cases}
1 & \text{if } u(I) \leq \tau, \\
0 & \text{if } u(I) > \tau,
\end{cases}
\end{equation}
where the threshold $\tau$ is set to $0.691$. 

This decision mechanism provides an intuitive reflection of the model's confidence: low uncertainty values are associated with authentic faces, whereas high uncertainty indicates a higher likelihood of forgery. Unlike conventional softmax-based probabilities, which often suffer from overconfidence, uncertainty quantification via EDL offers more reliable confidence estimation. This is particularly advantageous when encountering OOD forgery techniques that were not present during the training phase.

% \begin{figure*}[!t]
% \centering
% \subfloat[]{\includegraphics[width=3.56in]{df40.pdf}%
% \label{fig_first_case}}
% \hfil
% \subfloat[]{\includegraphics[width=3.56in]{asfd.pdf}%
% \label{fig_second_case}}
% \caption{Accuracy comparison between our FADNet and SOTA baselines on (a) DF40. and (b) ASFD.}
% \label{fig_sim}
% \end{figure*}
\begin{figure*}
    \centering
    \includegraphics[width=\linewidth]{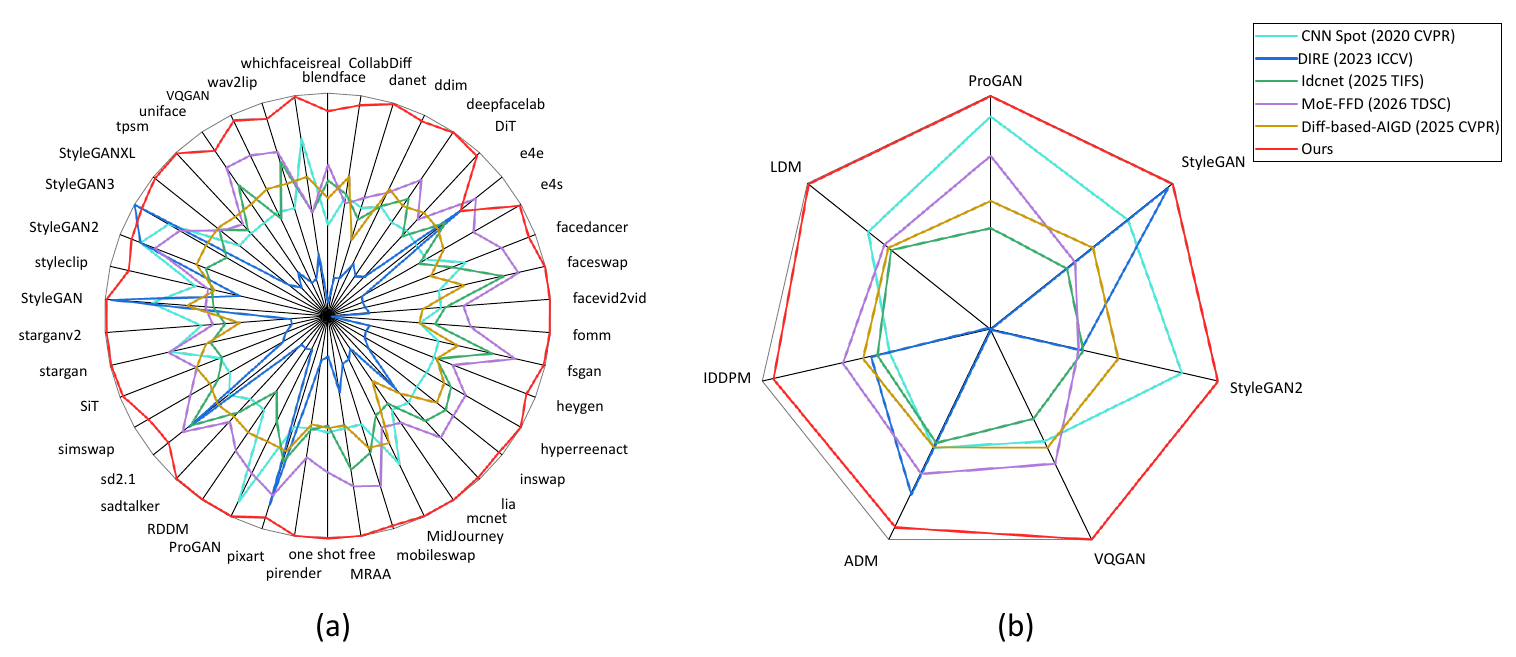}
    \caption{Accuracy comparison between our FADNet and SOTA baselines on (a) DF40. and (b) ASFD. It is evident that FADNet occupies the most extensive region within the radar plot, demonstrating its superior overall performance profile across all evaluated metrics relative to the baselines.  }
    \label{fig_sim}
\end{figure*}

\section{Experiment}
\label{sec:sec4}
This section provides a comprehensive experimental evaluation of the proposed FADNet. Firstly, we assess the generalization capability of our method by comparing it against five SOTA baseline methods on two large-scale datasets. Together, these benchmarks encompass over 40 diverse face forgery and manipulation techniques. Subsequently, we provide qualitative visualizations and conduct a detailed analysis of the uncertainty estimation produced by our model. Finally, an ablation study is performed to validate the rationality of our decision threshold selection strategy.

\subsection{Experimental Setup}
\textbf{Training Details.} Our FADNet is optimized using the AdamW optimizer with an initial learning rate of $5\times10^{-5}$ and a weight decay of $5\times10^{-4}$. We employ a cosine annealing scheduler to adjust the learning rate over 100 training epochs, with a batch size of 32. The training dataset comprises 10,000 real faces sourced from the FFHQ dataset. 
% To facilitate self-supervised one-class learning, 
% Meanwhile, we utilize a PFIG to produce a corresponding set of 10,000 pseudo-forgery faces, resulting in a balanced training distribution.  
Notably, although the real faces used for training and evaluation are both drawn from FFHQ, the two subsets are completely non-overlapping. Meanwhile, we utilize PFIG to pre-generate 10,000 pseudo-forgery faces before training, thereby establishing a balanced training distribution. Therefore, FADNet does not require any real forged images, while its training process involves both real samples and pseudo-forgery samples. All experiments are conducted on a single NVIDIA RTX5060Ti GPU.

\textbf{Datasets.} We conduct an extensive evaluation on two prominent benchmark datasets: DF40\footnote{\url{https://github.com/YZY-stack/DF40}}~\cite{NEURIPS2024_34239f60} and the Asian Synthetic Face Dataset\footnote{\url{https://github.com/Hurrice-star/ASFD}}~\cite{11209901} (ASFD). DF40 is a large-scale dataset encompassing 40 distinct forgery techniques. It provides a comprehensive spectrum of synthetic samples, ranging from traditional FaceSwap manipulations to cutting-edge GANs and DMs. ASFD addresses the representation gap in Asian facial features within existing benchmarks. Constructed using various generative models, the images in ASFD exhibit high visual fidelity and consistency with authentic imagery, thereby imposing more rigorous challenges on the discriminative boundaries of detection algorithms.

\textbf{Metrics.} To provide a holistic assessment of the proposed framework, the following metrics are chosen. We utilize Accuracy (Acc), F1-score, and Average Precision (AP) to evaluate the model's discriminative capability. Meanwhile, we adopt Expected Calibration Error~\cite{pmlr-v70-guo17a} (ECE) to quantify the alignment between the model's predictive confidence and its actual empirical accuracy. %By partitioning the confidence interval into $B=10$ equal-width bins and calculating the weighted average difference between accuracy and confidence, we assess the reliability of the model's probabilistic outputs. A lower ECE value indicates superior calibration, reflecting a more trustworthy detector. The ECE is defined as follows:$$\text{ECE} = \sum_{b=1}^{B} \frac{|B_b|}{N} \left| \text{acc}(B_b) - \text{conf}(B_b) \right|$$

\subsection{Compare with Existing Detectors}
\begin{table*}[!h]
\centering
\caption{Comprehensive detection performance on the DF40 dataset across all 40 generation methods.}
\label{tab:df40}
\resizebox{\textwidth}{!}{
\begin{tabular}{l|cc|cc|cc|cc|cc|cc}
\toprule
\multirow{2}{*}{\textbf{DF40}} & \multicolumn{6}{c|}{\textbf{Synthetic Detector}} & \multicolumn{4}{c|}{\textbf{Deepfake Detector}} & \multicolumn{2}{c}{} \\
\cmidrule(lr){2-7} \cmidrule(lr){8-11} \cmidrule(lr){12-13}
& \multicolumn{2}{c|}{\textbf{CNN Spot (2020)}} & \multicolumn{2}{c|}{\textbf{DIRE (2023)}} & \multicolumn{2}{c|}{\textbf{Diff-based (2025)}} & \multicolumn{2}{c|}{\textbf{IDCNet (2025)}} & \multicolumn{2}{c|}{\textbf{MoE-FFD (2026)}} & \multicolumn{2}{c}{\textbf{Ours}} \\
\cmidrule(lr){2-3} \cmidrule(lr){4-5} \cmidrule(lr){6-7} \cmidrule(lr){8-9} \cmidrule(lr){10-11} \cmidrule(lr){12-13}
& Acc / AP & F1 / ECE & Acc / AP & F1 / ECE & Acc / AP & F1 / ECE & Acc / AP & F1 / ECE & Acc / AP & F1 / ECE & Acc / AP & F1 / ECE \\
\midrule
blendface & 40.9 / 35.1 & 0.0 / 0.591 & 6.0 / 7.7 & 11.3 / 0.904 & 52.9 / 62.7 & 20.0 / 0.143 & 60.9 / 53.3 & 62.1 / 0.094 & 67.9 / 69.4 & 67.2 / 0.214 & \textbf{92.0} / \textbf{96.5} & \textbf{92.0} / 0.231 \\
CollabDiff & 45.4 / 46.0 & 14.2 / 0.506 & 0.9 / 1.3 & 1.8 / 0.987 & 55.9 / \textbf{100.0} & 21.1 / 0.125 & 46.2 / 46.3 & 39.1 / 0.236 & 41.4 / 37.3 & 17.9 / 0.516 & \textbf{94.8} / 97.8 & \textbf{94.8} / 0.217 \\
danet & 40.9 / 33.3 & 0.0 / 0.591 & 1.5 / 1.7 & 3.0 / 0.980 & 23.1 / 34.7 & 13.3 / 0.457 & 34.4 / 37.1 & 14.4 / 0.360 & 46.7 / 45.7 & 30.5 / 0.430 & \textbf{99.5} / \textbf{100.0} & \textbf{99.5} / 0.192 \\
ddim & 42.5 / 39.3 & 5.3 / 0.549 & 11.1 / 12.0 & 20.0 / 0.862 & 55.9 / \textbf{99.1} & 21.1 / 0.120 & 45.9 / 45.1 & 38.6 / 0.239 & 53.8 / 54.7 & 44.9 / 0.368 & \textbf{96.4} / \textbf{99.1} & \textbf{96.4} / 0.208 \\
deepfacelab & 40.9 / 32.2 & 0.0 / 0.591 & 6.0 / 7.1 & 11.3 / 0.910 & 51.7 / 51.5 & 19.6 / 0.208 & 56.5 / 54.4 & 55.9 / 0.129 & 68.9 / 66.2 & 68.6 / 0.197 & \textbf{99.5} / \textbf{100.0} & \textbf{99.5} / 0.192 \\
DiT & 41.0 / 35.7 & 0.3 / 0.588 & 9.0 / 9.1 & 16.5 / 0.896 & 55.9 / 99.3 & 21.1 / 0.122 & 38.8 / 39.1 & 24.4 / 0.313 & 50.9 / 50.4 & 39.3 / 0.377 & \textbf{97.6} / \textbf{99.4} & \textbf{97.6} / 0.201 \\
e4e & 59.3 / 63.0 & 47.5 / 0.348 & 71.5 / 68.3 & \textbf{83.4} / 0.163 & 55.3 / 58.6 & 20.9 / 0.253 & 60.5 / 56.7 & 61.5 / 0.094 & \textbf{81.5} / \textbf{89.0} & \textbf{83.4} / 0.127 & 70.8 / 80.7 & 70.3 / 0.450 \\
e4s & 40.9 / 32.9 & 0.0 / 0.591 & 0.9 / 0.8 & 1.8 / 0.991 & 51.6 / 52.5 & 19.6 / 0.190 & 36.6 / 37.8 & 19.5 / 0.335 & 70.3 / 74.3 & 70.4 / 0.212 & \textbf{98.9} / \textbf{99.9} & \textbf{98.9} / 0.192 \\
facedancer & 40.9 / 33.5 & 0.0 / 0.591 & 0.0 / 0.3 & 0.0 / 0.997 & 39.3 / 43.8 & 16.3 / 0.276 & 51.9 / 49.8 & 48.9 / 0.179 & 80.2 / 84.8 & 82.0 / 0.115 & \textbf{95.8} / \textbf{99.1} & \textbf{95.8} / 0.210 \\
faceswap & 40.9 / 34.6 & 0.0 / 0.591 & 0.6 / 0.9 & 1.2 / 0.989 & 55.1 / 66.5 & 20.8 / 0.119 & 76.7 / 81.7 & 80.4 / 0.098 & 84.8 / 99.0 & 86.8 / 0.159 & \textbf{99.5} / \textbf{100.0} & \textbf{99.5} / 0.192 \\
facevid2vid & 40.9 / 33.2 & 0.0 / 0.591 & 2.4 / 2.7 & 4.7 / 0.969 & 31.0 / 36.1 & 14.6 / 0.409 & 43.0 / 43.3 & 33.1 / 0.266 & 52.9 / 52.3 & 43.2 / 0.359 & \textbf{99.5} / \textbf{100.0} & \textbf{99.5} / 0.192 \\
fomm & 41.1 / 39.7 & 0.7 / 0.572 & 1.5 / 1.9 & 3.0 / 0.978 & 40.9 / 42.9 & 16.6 / 0.263 & 47.9 / 46.5 & 42.2 / 0.213 & 64.0 / 63.1 & 61.7 / 0.241 & \textbf{99.5} / \textbf{100.0} & \textbf{99.5} / 0.192 \\
fsgan & 40.9 / 33.5 & 0.0 / 0.587 & 3.0 / 4.4 & 5.8 / 0.941 & 51.4 / 56.5 & 19.5 / 0.138 & 70.5 / 68.7 & 73.8 / 0.060 & 82.9 / 95.7 & 84.9 / 0.158 & \textbf{98.7} / \textbf{99.8} & \textbf{98.7} / 0.195 \\
heygen & 41.0 / 31.9 & 0.3 / 0.590 & 1.8 / 2.4 & 3.5 / 0.973 & 42.7 / 39.7 & 17.1 / 0.402 & 42.5 / 42.7 & 32.1 / 0.291 & 51.7 / 54.8 & 40.9 / 0.412 & \textbf{94.4} / \textbf{99.0} & \textbf{94.4} / 0.203 \\
hyperreenact & 41.0 / 38.2 & 0.3 / 0.588 & 2.7 / 3.7 & 5.3 / 0.953 & 51.9 / 55.1 & 19.7 / 0.145 & 56.3 / 51.6 & 55.6 / 0.135 & 65.2 / 66.2 & 63.5 / 0.235 & \textbf{99.5} / \textbf{100.0} & \textbf{99.5} / 0.192 \\
inswap & 40.9 / 34.1 & 0.0 / 0.587 & 7.5 / 7.6 & 14.0 / 0.903 & 54.6 / 66.6 & 20.6 / 0.117 & 60.8 / 55.1 & 61.9 / 0.090 & 65.9 / 64.7 & 64.4 / 0.231 & \textbf{97.4} / \textbf{99.5} & \textbf{97.4} / 0.202 \\
lia & 43.4 / 38.6 & 8.1 / 0.538 & 33.3 / 33.5 & 50.0 / 0.570 & 36.6 / 42.1 & 15.7 / 0.293 & 56.8 / 54.9 & 56.4 / 0.131 & 69.0 / 72.1 & 68.7 / 0.216 & \textbf{98.4} / \textbf{99.8} & \textbf{98.4} / 0.195 \\
mcnet & 40.9 / 34.1 & 0.0 / 0.591 & 1.2 / 1.6 & 2.4 / 0.984 & 22.2 / 34.4 & 13.2 / 0.458 & 36.9 / 38.8 & 20.2 / 0.327 & 49.3 / 49.0 & 36.1 / 0.398 & \textbf{99.5} / \textbf{100.0} & \textbf{99.5} / 0.192 \\
MidJourney & 68.2 / 74.3 & 63.2 / 0.267 & 5.7 / 6.3 & 10.8 /  0.921 & 55.6 / 98.5 & 21.1 / 0.127 & 39.1 / 40.8 & 25.1 / 0.304 & 46.4 / 44.0 & 29.8 / 0.457 & \textbf{99.2} / \textbf{99.9} & \textbf{99.2} / 0.190 \\
mobileswap & 40.9 / 33.1 & 0.0 / 0.591 & 6.7 / 7.5 & 12.6 / 0.908 & 54.1 / 65.8 & 20.5 / 0.100 & 56.9 / 54.4 & 56.5 / 0.129 & 75.6 / 77.4 & 76.9 / 0.158 & \textbf{97.8} / \textbf{99.4} & \textbf{97.8} / 0.199 \\
MRAA & 40.9 / 34.2 & 0.0 / 0.587 & 21.3 / 21.9 & 35.2 / 0.700 & 39.4 / 42.8 & 16.3 / 0.266 & 63.3 / 56.2 & 65.2 / 0.081 & 72.6 / 73.6 & 73.3 / 0.176 & \textbf{99.5} / \textbf{100.0} & \textbf{99.5} / 0.192 \\
one\_shot\_free & 42.6 / 44.3 & 5.6 / 0.535 & 1.8 / 2.3 & 3.5 / 0.973 & 40.2 / 43.9 & 16.5 / 0.242 & 38.9 / 40.4 & 24.7 / 0.311 & 64.1 / 65.0 & 61.9 / 0.245 & \textbf{99.5} / \textbf{100.0} & \textbf{99.5} / 0.192 \\
pirender & 41.0 / 38.4 & 0.3 / 0.577 & 3.9 / 4.8 & 7.5 / 0.937 & 38.8 / 42.6 & 16.2 / 0.269 & 41.8 / 41.5 & 30.7 / 0.277 & 56.8 / 54.2 & 50.2 / 0.308 & \textbf{99.5} / \textbf{100.0} & \textbf{99.5} / 0.192 \\
pixart & 42.3 / 41.4 & 4.6 / 0.556 & 86.2 / 84.7 & 92.6 / 0.084 & 55.9 / \textbf{99.9} & 21.1 / 0.125 & 62.0 / 56.1 & 63.5 / 0.082 & 80.8 / 90.6 & 82.7 / 0.140 & \textbf{93.3} / 98.7 & \textbf{93.3} / 0.217 \\
ProGAN & 90.9 / 99.8 & 91.7 / 0.094 & 0.0 / 0.0 & 0.0 / 1.000 & 54.8 / 66.8 & 20.7 / 0.111 & 43.3 / 43.8 & 33.7 / 0.266 & 74.1 / 82.5 & 75.1 / 0.186 & \textbf{99.5} / \textbf{100.0} & \textbf{99.5} / 0.178 \\
RDDM & 40.9 / 31.4 & 0.0 / 0.591 & 1.2 / 1.3 & 2.4 / 0.986 & 55.9 / \textbf{99.5} & 21.1 / 0.121 & 28.9 / 44.2 & 0.0 / 0.384 & 67.9 / 64.1 & 67.2 / 0.204 & \textbf{99.4} / \textbf{100.0} & \textbf{99.4} / 0.192 \\
sadtalker & 40.9 / 33.0 & 0.0 / 0.591 & 0.9 / 1.7 & 1.8 / 0.977 & 54.0 / 60.9 & 20.4 / 0.149 & 50.3 / 47.1 & 46.3 / 0.193 & 57.6 / 59.5 & 51.6 / 0.302 & \textbf{99.5} / \textbf{100.0} & \textbf{99.5} / 0.192 \\
sd2.1 & 48.1 / 49.5 & 21.7 / 0.473 & 75.7 / 72.9 & 86.2 / 0.204 & 55.9 / \textbf{99.2} & 21.1 / 0.120 & 74.4 / 71.3 & 78.0 / 0.081 & 79.7 / 89.1 & 81.5 / 0.154 & \textbf{89.4} / 96.5 & \textbf{89.4} / 0.265 \\
simswap & 41.0 / 36.7 & 0.3 / 0.587 & 8.4 / 10.7 & 15.5 / 0.866 & 53.2 / 64.9 & 20.1 / 0.125 & 48.3 / 45.7 & 42.9 / 0.212 & 65.6 / 67.6 & 64.0 / 0.238 & \textbf{91.2} / \textbf{97.4} & \textbf{91.2} / 0.236 \\
SiT & 42.0 / 37.0 & 3.7 / 0.565 & 4.5 / 6.0 & 8.6 / 0.933 & 55.9 / \textbf{99.6} & 21.1 / 0.124 & 41.2 / 42.1 & 29.5 / 0.287 & 55.7 / 54.9 & 48.3 / 0.332 & \textbf{98.2} / \textbf{99.6} & \textbf{98.2} / 0.197 \\
stargan & 66.5 / 70.1 & 60.5 / 0.272 & 0.0 / 0.1 & 0.0 / 0.999 & 47.4 / 46.4 & 18.3 / 0.316 & 44.7 / 43.7 & 36.4 / 0.250 & 67.9 / 75.9 & 67.2 / 0.242 & \textbf{99.5} / \textbf{100.0} & \textbf{99.5} / 0.192 \\
starganv2 & 48.2 / 50.7 & 22.0 / 0.461 & 0.0 / 0.0 & 0.0 / 1.000 & 27.6 / 34.7 & 14.0 / 0.511 & 35.4 / 36.5 & 16.8 / 0.366 & 42.1 / 37.3 & 19.7 / 0.523 & \textbf{99.3} / \textbf{100.0} & \textbf{99.3} / 0.190 \\
StyleGAN & 75.0 / 81.5 & 73.2 / 0.200 & 97.1 / \textbf{100.0} & 49.3 / 0.015 & 55.9 / 92.6 & 21.1 / 0.103 & 41.7 / 42.3 & 30.5 / 0.285 & 46.1 / 43.8 & 29.2 / 0.455 & \textbf{99.5} / \textbf{100.0} & \textbf{99.5} / 0.178 \\
styleclip & 53.1 / 55.9 & 34.2 / 0.417 & 28.7 / 29.4 & 44.6 / 0.648 & 42.5 / 40.0 & 17.0 / 0.445 & 44.1 / 44.4 & 35.2 / 0.270 & 46.3 / 44.3 & 29.6 / 0.466 & \textbf{89.9} / \textbf{95.7} & \textbf{89.9} / 0.238 \\
StyleGAN2 & 88.2 / 94.4 & 88.9 / 0.085 & 88.3 / 86.9 & \textbf{93.8} / 0.073 & 55.9 / \textbf{100.0} & 21.1 / 0.125 & 50.5 / 48.6 & 46.6 / 0.194 & 79.9 / 89.4 & 81.7 / 0.143 & \textbf{93.5} / 98.6 & 93.5 / 0.220 \\
StyleGAN3 & 79.0 / 85.2 & 78.4 / 0.159 & \textbf{100.0} / \textbf{99.8} & \textbf{100.0} / 0.002 & 55.9 / \textbf{99.9} & 21.1 / 0.125 & 43.0 / 41.5 & 33.1 / 0.274 & 72.0 / 74.8 & 72.6 / 0.175 & 95.6 / 99.1 & 95.6 / 0.211 \\
StyleGANXL & 40.9 / 35.3 & 0.0 / 0.581 & 7.2 / 7.6 & 13.5 / 0.916 & 55.9 / 97.8 & 21.1 / 0.120 & 54.8 / 51.1 & 53.4 / 0.151 & 53.9 / 55.4 & 45.1 / 0.360 & \textbf{99.0} / \textbf{99.8} & \textbf{99.0} / 0.192 \\
tpsm & 40.9 / 32.6 & 0.0 / 0.591 & 0.9 / 1.5 & 1.8 / 0.983 & 53.2 / 56.5 & 20.1 / 0.148 & 43.2 / 42.4 & 33.5 / 0.265 & 47.3 / 46.6 & 31.8 / 0.420 & \textbf{99.5} / \textbf{100.0} & \textbf{99.5} / 0.192 \\
uniface & 40.9 / 35.8 & 0.0 / 0.589 & 8.4 / 11.1 & 15.5 / 0.866 & 53.2 / 64.1 & 20.1 / 0.121 & 65.6 / 59.9 & 68.1 / 0.068 & 76.8 / 80.7 & 78.3 / 0.134 & \textbf{87.9} / \textbf{94.3} & \textbf{87.9} / 0.269 \\
VQGAN & 42.1 / 41.6 & 4.0 / 0.555 & 0.0 / 0.4 & 0.0 / 0.996 & 55.9 / \textbf{100.0} & 21.1 / 0.125 & 38.3 / 39.3 & 23.3 / 0.315 & 76.1 / 83.0 & 77.5 / 0.154 & \textbf{96.9} / 99.2 & \textbf{96.9} / 0.203 \\
wav2lip & 40.9 / 33.8 & 0.0 / 0.591 & 1.2 / 1.5 & 2.4 / 0.984 & 54.7 / 67.9 & 20.7 / 0.116 & 67.0 / 58.0 & 69.8 / 0.098 & 72.5 / 72.4 & 73.2 / 0.168 & \textbf{91.1} / \textbf{97.4} & \textbf{91.1} / 0.253 \\
whichfaceisreal & 76.5 / 84.6 & 75.2 / 0.185 & 13.8 / 15.4 & 24.3 / 0.805 & 55.9 / 81.9 & 21.1 / 0.097 & 36.6 / 38.3 & 19.5 / 0.345 & 36.5 / 32.9 & 4.5 / 0.589 & \textbf{99.5} / \textbf{100.0} & \textbf{99.5} / 0.192 \\
\bottomrule
\end{tabular}}
\end{table*}

% % \textbf{Baselines.} To evaluate the performance of our proposed framework, we benchmark it against five representative SOTA baselines. These methods span several premier conferences and journals. They are, CNN Spot\footnote{\url{https://peterwang512.github.io/CNNDetection/}}~\cite{wang2020cnn}, DIRE\footnote{\url{https://github.com/ZhendongWang6/DIRE}}~\cite{wang2023dire}, IDCNet\footnote{\url{https://github.com/wangzhiyuan120/idcnet}}~\cite{11098842}, MoE-FFD\footnote{\url{https://github.com/LoveSiameseCat/MoE-FFD}}~\cite{11145290}, and Diff-based-AIGD\footnote{\url{https://github.com/Ekko-zn/Diff-based-AIGD}}~\cite{zhong2025beyond}.

\textbf{Baselines.} To evaluate the performance of our proposed framework, we benchmark it against five representative SOTA baselines. These methods span several premier conferences and journals. They are CNN Spot~\cite{wang2020cnn}, DIRE~\cite{wang2023dire}, IDCNet~\cite{11098842}, MoE-FFD~\cite{11145290}, and Diff-based~\cite{zhong2025beyond}.

\begin{table*}[htbp]
\centering
\caption{Detection performance on the ASFD dataset.}
\label{tab:asfd_results}
\resizebox{\textwidth}{!}{
\begin{tabular}{l|cc|cc|cc|cc|cc|cc}
\toprule
\multirow{2}{*}{\textbf{ASFD}} & \multicolumn{6}{c|}{\textbf{Synthetic Detector}} & \multicolumn{4}{c|}{\textbf{Deepfake Detector}} & \multicolumn{2}{c}{} \\
\cmidrule(lr){2-7} \cmidrule(lr){8-11} \cmidrule(lr){12-13}
& \multicolumn{2}{c|}{\textbf{CNN Spot (2020)}} & \multicolumn{2}{c|}{\textbf{DIRE (2023)}} & \multicolumn{2}{c|}{\textbf{Diff-based (2025)}} & \multicolumn{2}{c|}{\textbf{IDCNet (2025)}} & \multicolumn{2}{c|}{\textbf{MoE-FFD (2026)}} & \multicolumn{2}{c}{\textbf{Ours}} \\
\cmidrule(lr){2-3} \cmidrule(lr){4-5} \cmidrule(lr){6-7} \cmidrule(lr){8-9} \cmidrule(lr){10-11} \cmidrule(lr){12-13}
& Acc / AP & F1 / ECE & Acc / AP & F1 / ECE & Acc / AP & F1 / ECE & Acc / AP & F1 / ECE & Acc / AP & F1 / ECE & Acc / AP & F1 / ECE \\
\midrule
ProGAN    & 90.9 / 99.8 & 91.7 / 0.094 & 0.0 / 0.0 & 0.0 / 1.000 & 54.8 / 66.8 & 20.7 / 0.111 & 43.3 / 43.8 & 33.7 / 0.266 & 74.1 / 82.5 & 75.1 / 0.186 & \textbf{99.5} / \textbf{100.0} & \textbf{99.5} / 0.178 \\
StyleGAN  & 75.0 / 81.5 & 73.2 / 0.200 & 97.1 / \textbf{100.0} & 49.3 / 0.015 & 55.9 / 92.6 & 21.1 / 0.103 & 41.7 / 42.3 & 30.5 / 0.285 & 46.1 / 43.8 & 29.2 / 0.455 & \textbf{99.5} / \textbf{100.0} & \textbf{99.5} / 0.178 \\
StyleGAN2 & 83.8 / 89.9 & 84.1 / 0.122 & 39.6 / 75.4 & 18.9 / 0.506 & 55.9 / 91.3 & 21.1 / 0.082 & 40.6 / 40.4 & 28.3 / 0.300 & 38.5 / 34.9 & 10.2 / 0.556 & \textbf{99.5} / \textbf{100.0} & \textbf{99.5} / 0.178 \\
VQGAN     & 52.9 / 55.1 & 33.8 / 0.423 & 0.2 / 3.2 & 0.1 / 0.984 & 55.9 / 99.3 & 21.1 / 0.126 & 42.2 / 42.7 & 31.5 / 0.280 & 63.5 / 70.1 & 61.0 / 0.280 & \textbf{99.4} / \textbf{99.9} & \textbf{99.4} / 0.177 \\
ADM       & 56.1 / 57.3 & 40.9 / 0.392 & 78.2 / 87.7 & 43.9 / 0.164 & 55.8 / \textbf{98.6} & 21.1 / 0.121 & 53.7 / 53.0 & 51.7 / 0.164 & 68.3 / 74.0 & 67.8 / 0.226 & \textbf{93.7} / 97.9 & \textbf{93.7} / 0.209 \\
IDDPM     & 44.3 / 38.7 & 10.9 / 0.526 & 52.4 / 53.9 & 22.9 / 0.417 & 55.9 / \textbf{99.4} & 21.1 / 0.121 & 49.6 / 59.1 & 45.1 / 0.186 & 64.8 / 72.0 & 62.9 / 0.253 & \textbf{95.1} / 98.9 & \textbf{95.1} / 0.199 \\
LDM       & 66.9 / 72.0 & 61.1 / 0.281 & 1.1 / 28.1 & 0.7 / 0.981 & 55.9 / \textbf{100.0} & 21.1 / 0.125 & 54.4 / 53.7 & 52.8 / 0.154 & 58.0 / 59.2 & 52.3 / 0.321 & \textbf{99.5} / \textbf{100.0} & \textbf{99.5} / 0.178 \\
\bottomrule
\end{tabular}}
\end{table*}

\begin{table*}[t]
    \centering
    \caption{Detection performance comparison across commercial models.}
    \label{tab:multi_generator_results}
    \resizebox{\textwidth}{!}{
    \begin{tabular}{l|cc|cc|cc|cc|cc|cc}
    \toprule
    \multirow{2}{*}{\textbf{Dataset}} & \multicolumn{6}{c|}{\textbf{Synthetic Detector}} & \multicolumn{4}{c|}{\textbf{Deepfake Detector}} & \multicolumn{2}{c}{} \\
    \cmidrule(lr){2-7} \cmidrule(lr){8-11} \cmidrule(lr){12-13}
    & \multicolumn{2}{c|}{\textbf{CNN Spot (2020)}} & \multicolumn{2}{c|}{\textbf{DIRE (2023)}} & \multicolumn{2}{c|}{\textbf{Diff-based (2025)}} & \multicolumn{2}{c|}{\textbf{IDCNet (2025)}} & \multicolumn{2}{c|}{\textbf{MoE-FFD (2026)}} & \multicolumn{2}{c}{\textbf{Ours}} \\
    \cmidrule(lr){2-3} \cmidrule(lr){4-5} \cmidrule(lr){6-7} \cmidrule(lr){8-9} \cmidrule(lr){10-11} \cmidrule(lr){12-13}
    & Acc / AP & F1 / ECE & Acc / AP & F1 / ECE & Acc / AP & F1 / ECE & Acc / AP & F1 / ECE & Acc / AP & F1 / ECE & Acc / AP & F1 / ECE \\
    \midrule
   DALL-E 2   & 37.8 / 35.5 & 38.1 / 0.588 & 0.0 / 100.0 & 0.0 / 0.999 & 29.5 / 96.4 & 1.6 / \textbf{0.152} & 57.0 / 57.7 & 28.5 / 0.305 & 68.6 / 49.7 & 54.6 / 0.228 & \textbf{93.4} / \textbf{98.6} & \textbf{93.4} / 0.221 \\
    IF         & 41.6 / 57.6 & 50.1 / 0.512 & 39.3 / \textbf{100.0} & 56.4 / 0.537 & 50.1 / 68.2 & 1.6 / \textbf{0.133} & 66.9 / 77.0 & 62.9 / 0.196 & \textbf{71.3} / 76.4 & \textbf{71.7} / 0.191 & 50.8 / 82.7 & 35.9 / 0.444 \\
    Midjourney & 91.0 / 52.6 & 58.0 / \textbf{0.083} & 0.0 / \textbf{100.0} & 0.0 / 0.981 & 17.3 / 97.0 & 1.6 / 0.208 & 57.8 / 62.4 & 18.5 / 0.384 & 60.3 / 12.0 & 9.2 / 0.120   & \textbf{99.5} / \textbf{100.0} & \textbf{99.5} / 0.200 \\
    \bottomrule
    \end{tabular}}
    \end{table*}

\begin{table*}[htbp]
\centering
\caption{Detection performance on the latest generative paradigm (FLUX.2).}
\label{tab:flux_results}
\resizebox{\textwidth}{!}{
\begin{tabular}{l|cc|cc|cc|cc|cc|cc}
\toprule
\multirow{2}{*}{\textbf{FLUX}} & \multicolumn{6}{c|}{\textbf{Synthetic Detector}} & \multicolumn{4}{c|}{\textbf{Deepfake Detector}} & \multicolumn{2}{c}{} \\
\cmidrule(lr){2-7} \cmidrule(lr){8-11} \cmidrule(lr){12-13}
& \multicolumn{2}{c|}{\textbf{CNN Spot (2020)}} & \multicolumn{2}{c|}{\textbf{DIRE (2023)}} & \multicolumn{2}{c|}{\textbf{Diff-based (2025)}} & \multicolumn{2}{c|}{\textbf{IDCNet (2025)}} & \multicolumn{2}{c|}{\textbf{MoE-FFD (2026)}} & \multicolumn{2}{c}{\textbf{Ours}} \\
\cmidrule(lr){2-3} \cmidrule(lr){4-5} \cmidrule(lr){6-7} \cmidrule(lr){8-9} \cmidrule(lr){10-11} \cmidrule(lr){12-13}
& Acc / AP & F1 / ECE & Acc / AP & F1 / ECE & Acc / AP & F1 / ECE & Acc / AP & F1 / ECE & Acc / AP & F1 / ECE & Acc / AP & F1 / ECE \\
\midrule
FLUX.2     & 54.1 / 18.9 & - / 0.459 & 0.0 / 0.0 & 0.0 / 1.000 & 50.2 / 43.4 & 1.6 / 0.372 & 29.8 / 33.7 & 2.5 / 0.440 & 36.8 / 32.4 & 5.9 / 0.595 & \textbf{97.1} / \textbf{99.0} & \textbf{97.1} / 0.186 \\
\bottomrule
\end{tabular}}
\end{table*}

We compare the proposed FADNet against five representative baseline methods. And the accuracy radar charts for the two datasets are illustrated in Fig.~\ref{fig_sim}.

\subsubsection{Detection Performance Comparison}
Table~\ref{tab:df40} presents a detailed performance breakdown on the DF40 dataset. The experimental results demonstrate that FADNet significantly outperforms competing methods across the vast majority of forgery types. Specifically, legacy detection methods such as CNN Spot exhibit clear limitations when confronted with diverse novel forgery techniques. While DIRE leverages diffusion reconstruction error to detect diffusion-generated imagery, it exhibits substantial instability on non-diffusion-based forgeries. For instance, its accuracy drops to a mere 6.0\% on blendface and deepfacelab, highlighting a deficiency in cross-domain generalization. Although recently proposed frameworks like IDCNet and MoE-FFD improve overall performance through sophisticated feature-learning mechanisms, they still fall short of FADNet across all primary metrics. Furthermore, Diff-based yields disappointing results on DF40 as it is optimized for general synthetic image detection rather than the specific nuances of synthetic faces. Benefiting from our uncertainty-aware one-class learning strategy, FADNet effectively captures the underlying distribution of authentic facial features, leading to good discriminative power against unseen forgeries. The data indicates that FADNet maintains an accuracy exceeding 90\% across nearly all test categories and achieves near-optimal performance on several classes, including faceswap, deepfacelab, and StyleGAN, underscoring its superiority as a universal forgery face detector.

Table~\ref{tab:asfd_results} further details the comparative results on the ASFD dataset. This dataset comprises high-quality synthetic faces that demand fine-grained feature extraction capabilities. The performance of baseline methods fluctuates significantly on this dataset, FADNet demonstrates remarkable stability, achieving an accuracy of over 93\% across all seven generation types. To assess the generalization capability of FADNet in real-world scenarios, we tested it on face images generated by commercial models\footnote{\url{https://github.com/purdue-m2/AI-Face-FairnessBench}}. As shown in Table~\ref{tab:multi_generator_results}, FADNet significantly outperforms the comparison methods on DALL-E 2 and Midjourney, demonstrating strong generalization to commercial image generators. Nevertheless, its performance on DeepFloyd IF leaves room for further improvement. 
%修改过

\subsubsection{Generalization to Emerging Generative Paradigms}
Furthermore, to evaluate the generalization of our approach to previously unseen generative paradigm, we benchmarked baseline methods on images synthesized by Flux.2\footnote{\url{https://playground.bfl.ai/image/generate}}. Distinct from DMs, Flux.2 is built upon Flow Matching (FM)---a cutting-edge framework that delivers synthesis quality surpassing that of DMs, thereby representing a formidable new paradigm in image generation. For evaluation, we generated 1,000 images using FLUX.2 with the uniform prompt “a human face” at a resolution of 512 × 512 pixels, while retaining the default settings for all other generation parameters.
% In our experiments, all test samples were generated using the uniform prompt: 'a human face.' 
As illustrated in Table \ref{tab:flux_results}, existing baseline methods exhibit a significant performance degradation when encountering images produced by this novel paradigm. In contrast, our proposed FADNet maintains consistently high performance, demonstrating superior generalization capabilities.

\subsubsection{Comparison of Expected Calibration Error}
In security-critical forensic applications, a model must not only provide accurate predictions but also output confidence levels that align with its actual performance. We employ the ECE to quantify the consistency between the model's predictive confidence and its empirical accuracy. A lower ECE indicates a more reliable confidence estimation. The statistical results in 
%修改过
Table~\ref{tab:df40}, Table~\ref{tab:asfd_results} and Table~\ref{tab:multi_generator_results} reveal that FADNet exhibits superior stability in calibration performance compared to existing baselines. Current methods suffer from high calibration errors, suggesting that their incorrect predictions are often accompanied by extreme overconfidence that is highly misleading in real-world forensic deployments. In contrast, FADNet consistently maintains ECE values within a low range across all subsets.

\begin{table}[htbp]
\centering
\caption{Comparison of inference time on both datasets.}
\label{tab:inference_time_horizontal}
\setlength{\tabcolsep}{2pt} 
\footnotesize
\begin{tabular}{c|cccccc}
\toprule
\textbf{Dataset} & CNN Spot & DIRE & Diff-based & IDCNet & MoE-FFD & \textbf{Ours} \\
\midrule
DF40 & \textbf{48.4s} & 195.1s & 286.7s & 1344.6s & 253.1s & 145.8s \\
ASFD & 22.2s & 68.2s & 21.5s & 254.6s & 50.8s & \textbf{20.9s} \\
\bottomrule
\end{tabular}
\end{table}
\begin{table}[htbp]
\centering
\caption{Performance comparison between SeeABLE and FADNet on different datasets.}
\label{tab:seeable_comparison}
\resizebox{0.48\textwidth}{!}{
\begin{tabular}{c|ccc|ccc}
\toprule
\multirow{2}{*}{\textbf{Dataset}}
& \multicolumn{3}{c|}{\textbf{SeeABLE}}
& \multicolumn{3}{c}{\textbf{Ours}} \\
\cmidrule(lr){2-4}\cmidrule(lr){5-7}
& Acc (\%) & F1 (\%) & AP (\%)
& Acc (\%) & F1 (\%) & AP (\%) \\
\midrule
DF40
& 66.39 & 79.78 & \textbf{99.95}
& \textbf{95.30} & \textbf{95.20} & 98.40 \\

ASFD
& 36.40 & 50.68 & 84.10
& \textbf{98.03} & \textbf{98.03} & \textbf{99.53} \\

Commercial
& 53.67 & 50.85 & 50.96
& \textbf{81.23} & \textbf{76.27} & \textbf{93.77} \\
\bottomrule
\end{tabular}}
\end{table}

\subsubsection{Inference Efficiency Comparison}
To evaluate the computational efficiency of FADNet, we conducted comparative experiments on inference speed against several baseline methods, as summarized in Table \ref{tab:inference_time_horizontal}. On the ASFD dataset, FADNet achieves the superior inference speed with a total time of 20.87s, outperforming all baseline models. Our approach reduces inference time by approximately 91.8\% compared to IDCNet, which exhibits the highest computational overhead. On the larger and more complex DF40 dataset, while CNN Spot maintains the fastest detection speed, FADNet remains highly competitive and significantly outperforms the majority of other methods.

\subsubsection{Comparison with SeeABLE}

To further evaluate the effectiveness of FADNet, we compared it with SeeABLE, which also trains its detector by constructing artificial discrepancies from real images. As presented in Table~\ref{tab:seeable_comparison}, we report the comparison of the average performance metrics between FADNet and SeeABLE across the DF40, ASFD, and commercial model datasets. FADNet consistently outperforms SeeABLE in terms of both Accuracy and F1-score across all three datasets. Remarkably, on ASFD, the Accuracy improves from 36.40\% to 98.03\%. Although SeeABLE achieves a slightly higher AP on DF40, its overall classification performance is substantially lower than that of FADNet. Detailed comparison results are provided in Section III of the supplementary material.

In summary, the extensive experimental results across both benchmarks validate that FADNet not only outperforms all existing SOTA baselines in standard classification metrics but, more importantly, demonstrates the generalization and reliability essential for detecting previously unseen and novel face forgeries.

\begin{figure}
    \centering
    \includegraphics[width=\linewidth]{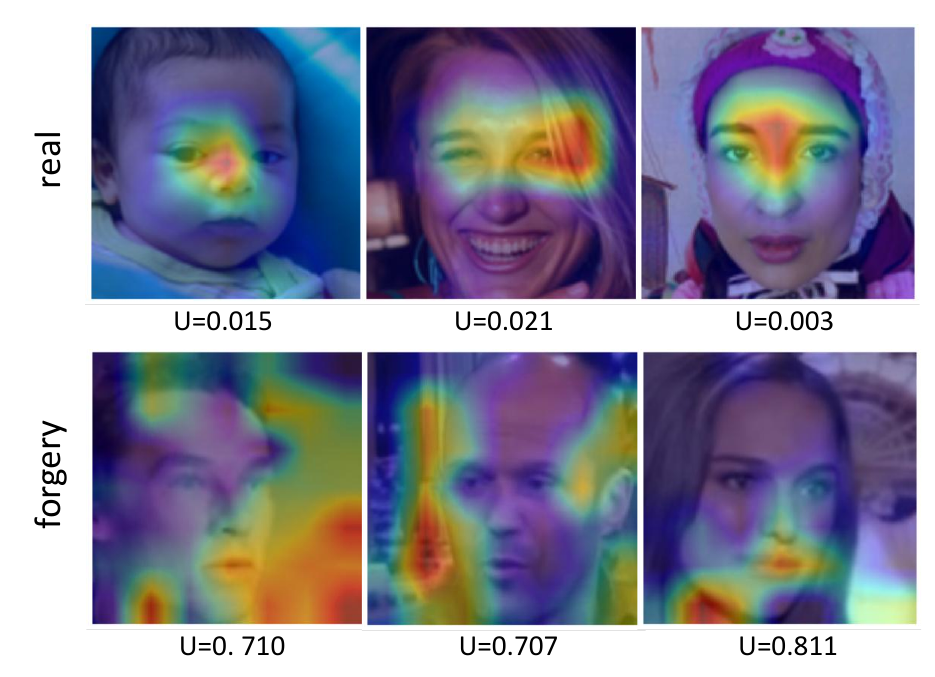}
    \caption{Comparative Grad-CAM visualizations of real versus forged faces. U represents uncertainty value. The top row displays the heatmaps of authentic faces, while the bottom row illustrates those of forged samples. It is evident that for real faces, the heatmaps are concentrated on core facial regions with a localized focus. In contrast, forged faces exhibit a more dispersed and irregular distribution. Furthermore, authentic faces yield lower uncertainty values.}
    \label{fig:grad}
\end{figure}
\begin{figure*}
    \centering
    \includegraphics[width=\linewidth]{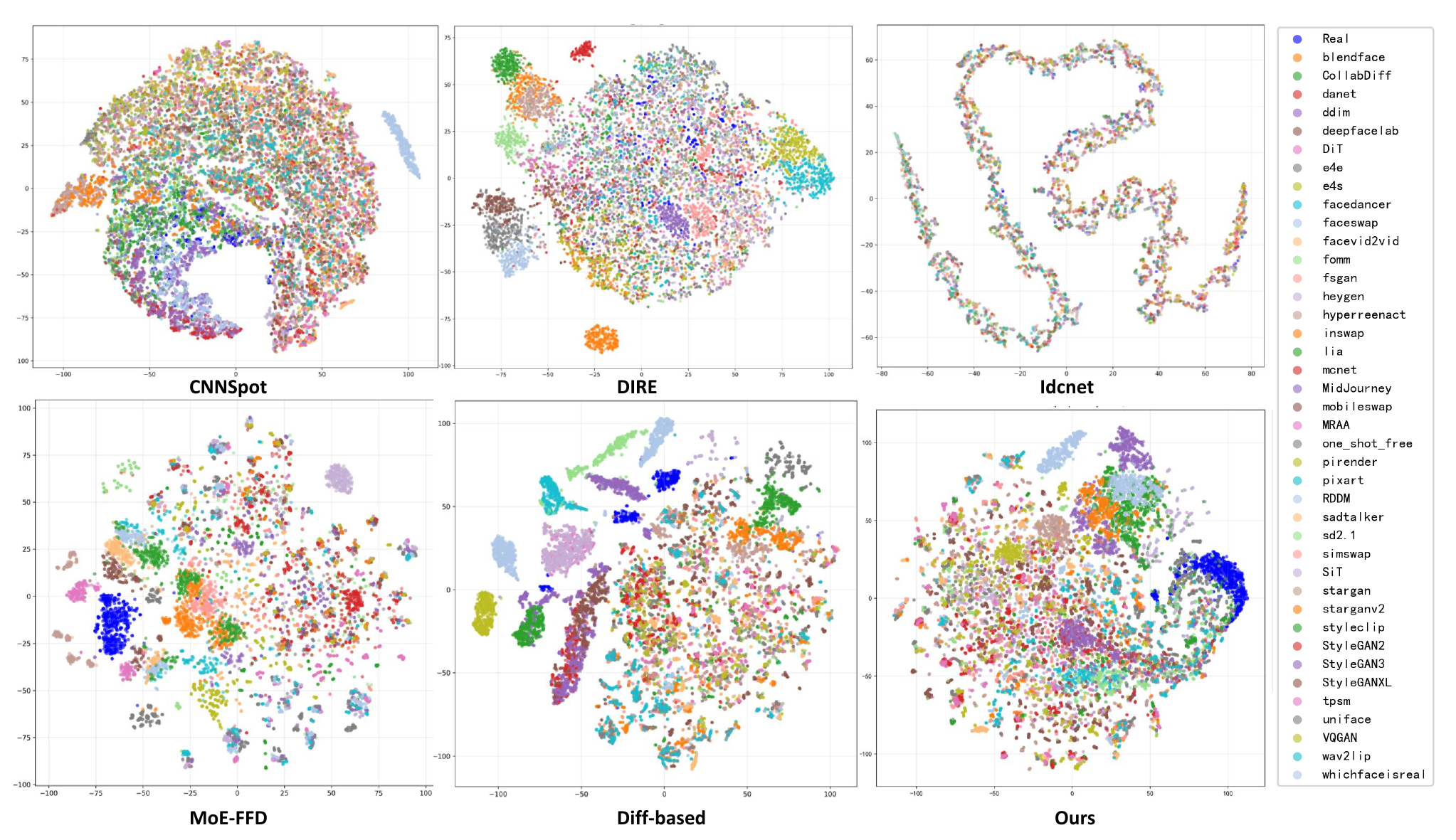}
    \caption{Comparative t-SNE analysis of feature distributions between FADNet and SOTA approaches. The blue clusters represent real samples, while other colors denote various forgery types. Notably, MoE-FFD also achieves tightly clustered representations for real samples, but its focus on distinguishing various forgery types results in a decision boundary overfitted to specific generator fingerprints. Conversely, FADNet captures the intrinsic distribution of real faces to form a tightly clustered representation. By treating forgeries as a unified deviation from this intrinsic manifold, our model mitigates the learning of source-specific artifacts and establishes a more robust decision boundary, explaining its superior generalization across diverse datasets.}
    \label{fig:alltsne}
\end{figure*}

\subsection{Visualizations}
To further validate the efficacy of FADNet’s uncertainty-aware mechanism, we conduct an analysis of its decision logic using Grad-CAM~\cite{selvaraju2017grad} visualizations. As illustrated in Fig.~\ref{fig:grad}, when processing authentic facial images, FADNet's attention is precisely localized on key biometric regions, yielding minimal uncertainty. In contrast, for forged images, the resulting activation maps exhibit a distinct diffusion pattern, with the model's focus extending toward facial boundaries and background regions, accompanied by a significant surge in predictive uncertainty. This difference effectively demonstrates that the uncertainty-aware mechanism in FADNet provides a more interpretable and physical grounding for complex classification decisions.

Additionally, we employ the t-SNE~\cite{maaten2008visualizing} visualization method to project the feature vectors extracted from the final layer of FADNet into a lower-dimensional space. As illustrated in Fig.~\ref{fig:alltsne}, authentic images are tightly clustered, whereas forged samples exhibit a dispersed distribution. This observation aligns with our expectations, demonstrating that FADNet has successfully characterized a discriminative feature space for authentic imagery.

\subsection{Ablation Study}
\subsubsection{Threshold selection}
\begin{figure}
    \centering
    \includegraphics[width=0.8\linewidth]{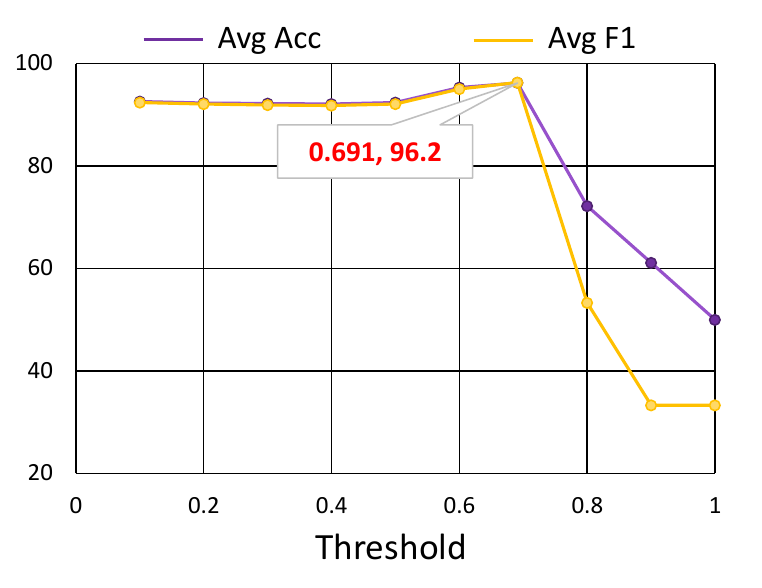}
    \caption{Ablation study of the decision threshold. The model achieves optimal accuracy and F1-score at a threshold of 0.691.}
    \label{fig:ablation}
\end{figure}

Since FADNet utilizes a decision threshold to distinguish between authentic and forged images, the selection of this threshold has a critical impact on classification performance. We selected the decision threshold on the DF40 validation set, which was disjoint from the final test set, by sweeping $\tau$ from $0.1$ to $1.0$ and evaluating the corresponding accuracy and F1-score, as illustrated in Fig.~\ref{fig:ablation}. Overall, the performance metrics exhibit a progressive improvement as the threshold increases from $0.1$ up to $0.691$, reaching an optimal peak at $\tau = 0.691$. However, beyond this point, a noticeable decline in performance occurs as the threshold continues to increase. This trend suggests that $\tau = 0.691$ represents the optimal decision boundary for balancing precision and recall within the uncertainty space. Accordingly, we consistently use this fixed threshold in all evaluations, including those conducted on the ASFD dataset and commercial models datasets.
\begin{table}[htbp]
  \centering
  \caption{Ablation Study of Loss Function}
  \label{tab:ablation_study}
  \begin{tabular}{lcc}
    \toprule
   Loss &Acc &F1-score \\
    \midrule
    Full           & \textbf{97.25} & \textbf{98.50} \\
    w/o Uncertainty Regularization  & 50.00          & 33.33          \\
    w/o Feature Compactness Loss   & 76.88          & 86.87          \\
    \bottomrule
  \end{tabular}
\end{table}

\begin{table*}[!h]
\centering
\caption{Ablation study on the DF40 dataset (Average results). This table compares the full model against versions with one component removed: Frequency Domain Perturbations (Freq.), Spatial and Structural Inconsistencies (S.S.), Chromatic and Textural Statistics (C.T.), and Compression (Comp.).}
\label{tab:ablation_average}
\resizebox{\textwidth}{!}{
\begin{tabular}{l|cc|cc|cc|cc|cc}
\toprule
DF40 & \multicolumn{2}{c|}{Ours (Full)} & \multicolumn{2}{c|}{w/o Freq.} & \multicolumn{2}{c|}{w/o S.S.} & \multicolumn{2}{c|}{w/o C.T.} & \multicolumn{2}{c}{w/o Comp.} \\
\cmidrule(lr){2-3} \cmidrule(lr){4-5} \cmidrule(lr){6-7} \cmidrule(lr){8-9} \cmidrule(lr){10-11}
& Acc / AP & F1 / ECE & Acc / AP & F1 / ECE & Acc / AP & F1 / ECE & Acc / AP & F1 / ECE & Acc / AP & F1 / ECE \\
\midrule
\textbf{Average} & \textbf{95.3 / 98.4} & \textbf{95.2 / 0.216} & \textbf{86.3 / 90.7} & \textbf{84.5 / 0.266} & \textbf{89.4 / 90.7} & \textbf{88.1 / 0.246} & \textbf{83.9 / 87.7} & \textbf{81.8 / 0.283} & \textbf{78.7 / 89.8} & \textbf{76.8 / 0.316} \\
\bottomrule
\end{tabular}}
\end{table*}
\begin{table}
\centering
\caption{Average performance comparison between ResNet-50 and DINOv2 on the DF40 dataset. }
\label{tab:ours_vs_dinov2_average}
\begin{tabular}{c|cc|cc}
\toprule
DF40 & \multicolumn{2}{c|}{ResNet-50} & \multicolumn{2}{c}{DINOv2} \\
\cmidrule(lr){2-3} \cmidrule(lr){4-5}
& Acc / AP & F1 / ECE & Acc / AP & F1 / ECE \\
\midrule
\textbf{Average} & \textbf{95.3 / 98.4} & \textbf{95.2 / 0.216} & 82.8 / 89.6 & 81.8 / 0.239 \\
\bottomrule
\end{tabular}%
\end{table}

\subsubsection{Ablation of Loss Components}
To evaluate the individual contribution of each loss component to the overall face forgery detection performance of FADNet, we conducted a systematic ablation study. Specifically, we compared the full model (Full) against variants where key loss terms were removed. The quantitative results are summarized in Table~\ref{tab:ablation_study}. The removal of the Uncertainty Regularization term results in a precipitous drop in Acc to 50.00\% and an F1-score of 33.33\%. This catastrophic performance degradation underscores that uncertainty modeling is the foundational pillar of FADNet, enabling the model to define a meaningful decision boundary in a one-class setting. Furthermore, omitting the Feature Compactness Loss leads to a decrease in accuracy to 76.88\% and an F1-score of 86.87\%. This decline demonstrates that the compactness constraint forces the feature distribution of authentic faces to be more cohesive, effectively increasing the margin between real samples and scattered forgery outliers. Consequently, this significantly bolsters the discriminative capacity of the decision boundary. The experiments reveal that while uncertainty regularization is indispensable for enabling facial forensics, feature compactness further refines and optimizes detection precision. The synergy between these components allows FADNet to effectively characterize the intrinsic distribution of authentic facial features.

\subsubsection{Contribution of Different Perturbations in PFIG}
To evaluate the individual contributions of different perturbation types to PFIG, we conducted a series of systematic ablation studies on the DF40 dataset. Table~\ref{tab:ablation_average} reports the average performance of our full model and its ablated variants, while results for each specific subset of the DF40 dataset are detailed in the supplementary material. Each ablated variant was trained with one of the four perturbation types removed. Overall, the full model achieves the best average performance across all metrics, indicating that the framework possesses not only high detection accuracy but also superior confidence calibration. The removal of any single perturbation type leads to varying degrees of performance degradation, underscoring the critical role each perturbation plays in the training of FADNet.

\subsubsection{Impact of Different Backbones}
To investigate the impact of different backbone networks on the performance of FADNet, we employed ResNet-50 and DINOv2 as feature extractors while keeping all other experimental settings unchanged, and compared their performance on the DF40 dataset. As shown in Table~\ref{tab:ours_vs_dinov2_average}, FADNet achieves an accuracy of 95.3\% with ResNet-50, substantially outperforming its DINOv2-based counterpart. ResNet-50 also yields a lower ECE, indicating better calibration of the model’s predictive confidence. Although DINOv2 offers strong semantic representations, it may be less sensitive to the subtle texture and frequency artifacts required for face forgery detection. In contrast, ResNet-50 better preserves these low-level cues and more accurately models the distribution of authentic faces. We therefore adopt ResNet-50 as the backbone of FADNet.

\begin{figure}
    \centering
    \includegraphics[width=\linewidth]{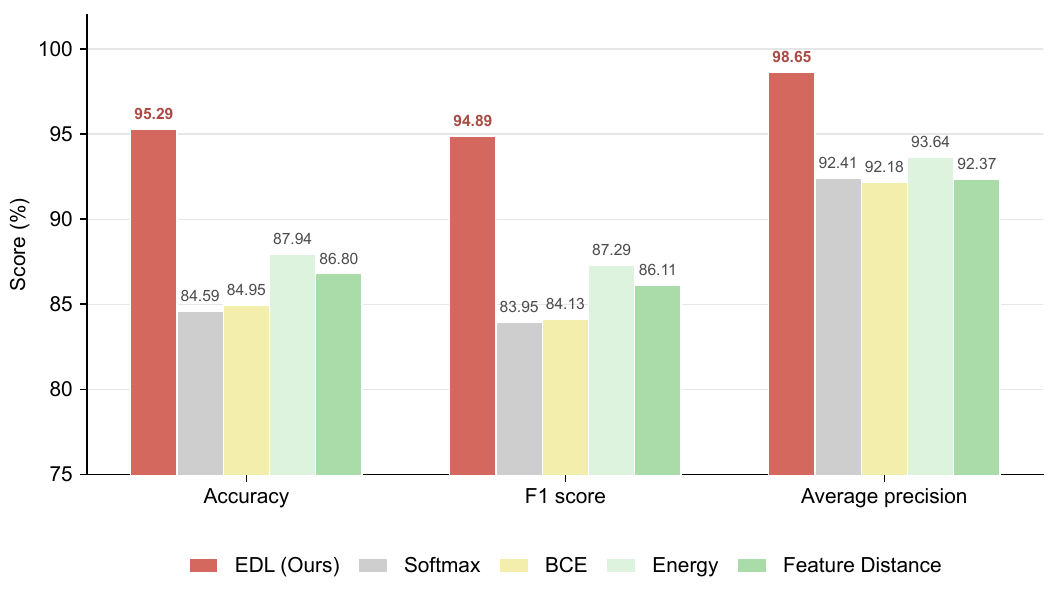}
    \caption{Performance comparison of different detection scoring strategies.}
    \label{fig:fig8}
\end{figure}
\subsubsection{Impact of Different Detection Scoring Strategies}
To verify whether the performance gain stems from evidential modeling itself, rather than merely from other  factors, we compared different detection scoring strategies under identical experimental conditions. We further implemented four representative baseline strategies: Softmax confidence-based scoring, BCE confidence-based scoring, energy-based scoring, and feature distance-based scoring. This design ensures that performance disparities across methods primarily reflect the impact of the detection scoring formulation. As shown in Fig.~\ref{fig:fig8}, EDL achieves the highest average Accuracy, F1-score, and AP on the DF40 dataset, outperforming all alternative scoring strategies. These results demonstrate that incorporating EDL indeed significantly improves the performance of our proposed framework.

\subsection{Robustness}
\begin{figure}
    \centering
    \includegraphics[width=\linewidth]{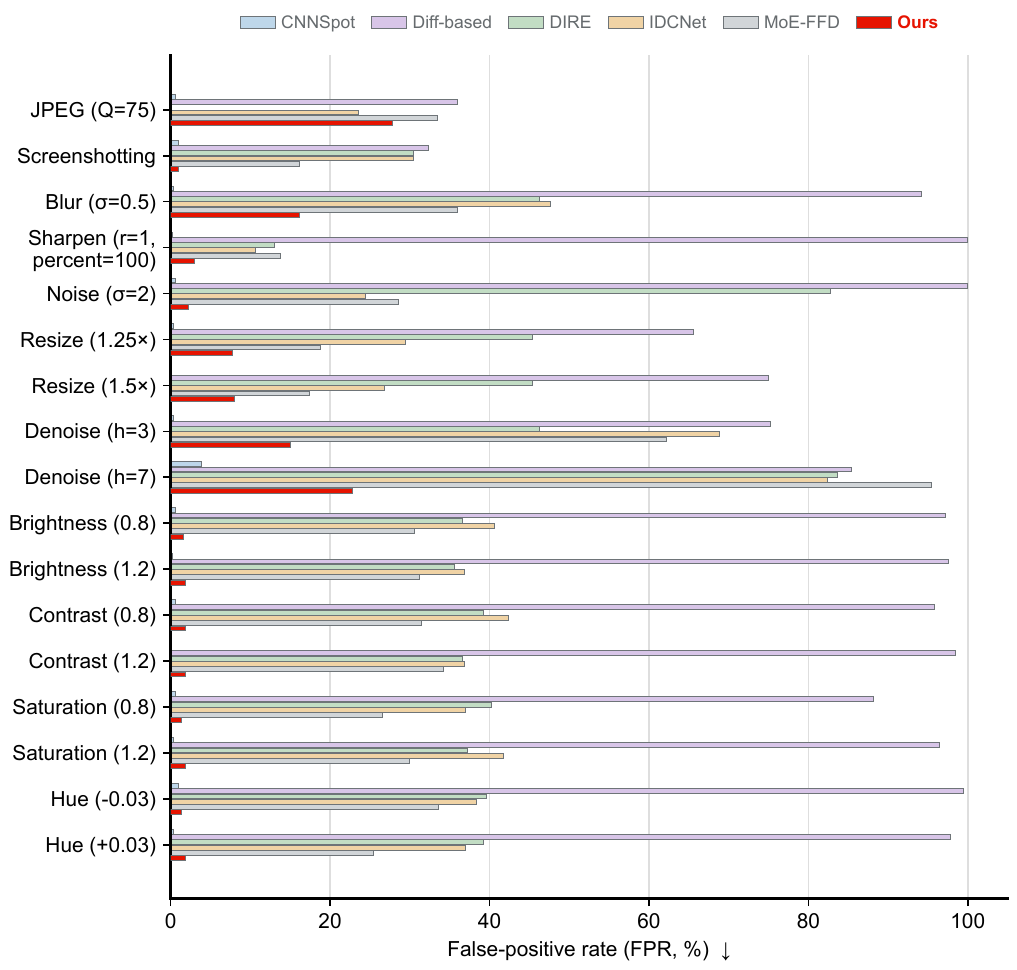}
    \caption{FPR comparison under several common post-processing operations.}
    \label{fig:fig9}
\end{figure}

During real-image acquisition, editing, and transmission, images are often inevitably subjected to various post-processing operations. To quantify the impact of these common operations on false detections, we evaluated the false-positive rate (FPR) under a range of post-processing conditions and compared FADNet with several baseline methods, as shown in Fig.~\ref{fig:fig9}. The results demonstrate that, compared with these baselines, FADNet maintains satisfactory robustness against various perturbations. In particular, under several evaluated scenarios, FADNet achieves consistently low false-positive rates, with very few real images being misclassified as forged. Extending the framework to maintain stable performance across a broader range of image transformations represents a promising direction for future work.
\section{Conclusion}
\label{sec:sec5}

In this paper, we propose FADNet, a one-class learning and uncertainty-aware method for face authenticity  detection. Departing from fully supervised paradigms that rely heavily on diverse forged samples for training, our approach leverages exclusively authentic facial data. By integrating pseudo-forgery feature simulation with evidential uncertainty modeling, FADNet achieves superior generalization capabilities against unseen forgery models, even in the face of entirely new paradigms. Extensive experimental results on several large-scale benchmarks demonstrate that FADNet significantly surpasses current SOTA methods in terms of detection accuracy, cross-dataset generalization, and confidence calibration. In the future, we will focus on exploring more efficient forgery detection architectures and further achieving tighter clustering of authentic images in the feature space.

%Furthermore, the uncertainty-driven decision mechanism provides enhanced reliability and interpretability for real-world digital forensic applications.
\bibliographystyle{IEEEtran}
\bibliography{refs}

\end{document}